Original Paper

# Using a Personal Health Library–Enabled mHealth Recommender System for Self-Management of Diabetes Among Underserved Populations: Use Case for Knowledge Graphs and Linked Data

Nariman Ammar[1], MSc, PhD; James E Bailey[2], MD, MPH; Robert L Davis[1], MD, MPH; Arash Shaban-Nejad[1], MSc, MPH, PhD

[1]Oak Ridge National Laboratory Center for Biomedical Informatics, Department of Pediatrics, College of Medicine, The University of Tennessee Health Science Center, Memphis, TN, United States

[2]Center for Health System Improvement, College of Medicine, The University of Tennessee Health Science Center, Memphis, TN, United States

**Corresponding Author:**
Arash Shaban-Nejad, MSc, MPH, PhD
Oak Ridge National Laboratory Center for Biomedical Informatics, Department of Pediatrics
College of Medicine
The University of Tennessee Health Science Center
50 N Dunlap Street
Memphis, TN, 38103
United States
Phone: 1 901 287 5863
Email: ashabann@uthsc.edu

## Abstract

**Background:** Traditionally, digital health data management has been based on electronic health record (EHR) systems and has been handled primarily by centralized health providers. New mechanisms are needed to give patients more control over their digital health data. Personal health libraries (PHLs) provide a single point of secure access to patients' digital health data and enable the integration of knowledge stored in their digital health profiles with other sources of global knowledge. PHLs can help empower caregivers and health care providers to make informed decisions about patients' health by understanding medical events in the context of their lives.

**Objective:** This paper reports the implementation of a mobile health digital intervention that incorporates both digital health data stored in patients' PHLs and other sources of contextual knowledge to deliver tailored recommendations for improving self-care behaviors in diabetic adults.

**Methods:** We conducted a thematic assessment of patient functional and nonfunctional requirements that are missing from current EHRs based on evidence from the literature. We used the results to identify the technologies needed to address those requirements. We describe the technological infrastructures used to construct, manage, and integrate the types of knowledge stored in the PHL. We leverage the Social Linked Data (Solid) platform to design a fully decentralized and privacy-aware platform that supports interoperability and care integration. We provided an initial prototype design of a PHL and drafted a use case scenario that involves four actors to demonstrate how the proposed prototype can be used to address user requirements, including the construction and management of the PHL and its utilization for developing a mobile app that queries the knowledge stored and integrated into the PHL in a private and fully decentralized manner to provide better recommendations.

**Results:** To showcase the main features of the mobile health app and the PHL, we mapped those features onto a framework comprising the user requirements identified in a use case scenario that features a preventive intervention from the diabetes self-management domain. Ongoing development of the app requires a formative evaluation study and a clinical trial to assess the impact of the digital intervention on patient-users. We provide synopses of both study protocols.

**Conclusions:** The proposed PHL helps patients and their caregivers take a central role in making decisions regarding their health and equips their health care providers with informatics tools that support the collection and interpretation of the collected knowledge. By exposing the PHL functionality as an open service, we foster the development of third-party applications or services and provide motivational technological support in several projects crossing different domains of interest.

*(JMIR Form Res 2021;5(3):e24738)* doi: 10.2196/24738

## *Introduction*

### Overview

Historically, medicine has been largely health care provider–centered rather than patient-centered [1-3]. However, the new trend is moving toward incorporating patients' social and behavioral characteristics into electronic health records (EHRs) [4]. This combination of medical, social, behavioral, and lifestyle information about the patient is essential to facilitate understanding of medical events in the context of one's life and, conversely, to allow lifestyle choices to be considered jointly with that patient's medical context [5]. This data is generated over time by patients, their caregivers, and their providers and is potentially useful to all parties for decision making [6]. Patients are increasingly frustrated by the lack of EHR interoperability among fragmented systems and platforms dictated by providers or insurers, and they have expressed their needs to have an active role in managing their own health care data [7-12]. Improved interoperability and support for patient-provider communication have the potential to improve patient satisfaction and, evidence suggests, could even help detect and prevent medical errors [12].

In the field of personal digital health management, we often distinguish between a health *state* and a health *process* [12]. Health *state* is a digital representation of the patients' health at a given point in time, including their prescribed and over-the-counter medications, test results, exercise regimens, diets, appointments with providers, and clinical outcomes. Patients also receive other digital health information through other communication modalities and from diverse sources. These data can often come in different formats depending on their source, including EHRs, family histories, data streams from activity trackers, published research documents and data sets, websites, social media platforms, and videos. A health state changes over time as data is acquired through ongoing processes and events embedded within those processes. A health-related intervention is a *process*, which could either be therapeutic or preventive. For instance, an intervention for self-management of diabetes mellitus (commonly referred to simply as diabetes) focused on promoting change or changes in health behavior to improve clinical outcomes is an example of a preventive intervention. Early interventions are the best way to prevent the progression of a negative health outcome to its end stage. Digital interventions through mobile Health (mHealth) apps can serve as an effective tool in chronic disease self-management [13]. An integrated *personal health library* (PHL) can facilitate the building of mHealth apps by maintaining a historical digital representation of a patient's health state from diagnosis to monitoring and integrating local knowledge about patients with global web-based knowledge while providing patients with full ownership of their digital health states. Applications can process the knowledge stored in the library to generate intelligent personalized interventions that can help improve patients' health state.

### Objective

We recently proposed both the conceptualization and initial implementation plan of a PHL [14,15]. The proposed PHL architecture (Figure 1) is the first of its kind to incorporate privacy, data ownership, integration, interoperability, portability, dynamic knowledge discovery, social determinants of health (SDoH) [16], and observations of daily living (ODLs) [5] into an end-to-end framework. In this paper, we provide a thematic assessment of patient requirements in a PHL, demonstrate how our proposed PHL meets these requirements, and describe an mHealth app that queries the PHL to deliver intelligent recommendations for improving self-care behaviors in diabetic adults.

**Figure 1.** A PHL that leverages the semantic technologies and decentralized privacy and security mechanisms of Social Linked Data (Solid) to enable true ownership, data integration, interoperability, portability, and dynamic knowledge discovery. The PHL enables building Hybrid mHealth Recommenders and Digital Librarians (HRDLs). ACL: access control list; API: application programming interface; Bp: blood pressure; DWPC: Diabetes Wellness and Prevention Coalition; ED: emergency department; LDN: Linked Data Notifications; LDP: Linked Data Platform; LOD: Linked Open Data; mHealth: mobile health; OWL: Web Ontology Language; PKG: personal knowledge graph; RDF: Resource Description Framework; REST: representational state transfer; SPARQL: SPARQL Protocol and RDF Query Language; WAC: Web Access Control.

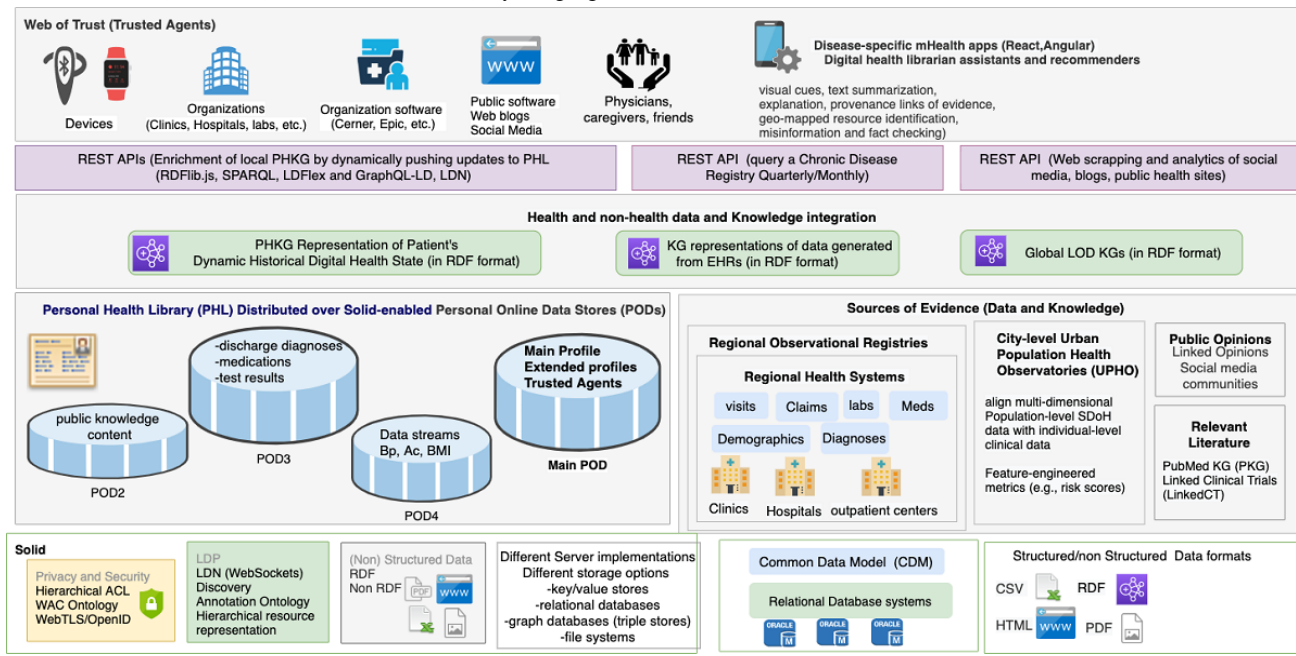

### Review of Relevant Literature

To the best of our knowledge, little data exist on the implementation of PHLs. Barr et al [17,18] proposed the Audio-PaHL project that utilizes text, audio, and image mining, natural language processing (NLP), and social network analysis to integrate audio-recordings of clinic visits into a library. They link medical terms that appear in the recordings to trustworthy patient resources, which can be retrieved, organized, edited, and shared by patients. Their project enables self-management in caregivers and older adults with multimorbidity. Several other researchers have utilized Semantic Web technologies and techniques to enhance EHRs or build research prototypes that can be integrated within clinical workflows. Table 1 describes some of the efforts in this area, including the rationale, the methods used, the health outcomes on which they focused, and whether they incorporate PHLs, SDoH, ODLs, and privacy. These works utilized a mix of Semantic Web and Machine Learning techniques. Most of these prior studies have predominantly focused on diabetes as the health outcome of interest. However, none of them incorporated privacy, SDoH, and ODLs in an integrated end-to-end framework.

Table 1. Comparison with existing methods.

| Reference | Rationale | Method | Health outcome | Utilizes a PHL[a] | Privacy-aware | Incorporates ODLs[b] | Incorporates SDoH[c] |
| --- | --- | --- | --- | --- | --- | --- | --- |
| Audio-PaHL Barr et al [17,18] | Self-management | Audio text retrieval; Natural Language Processing (NLP); Social Network Analysis | Multiple | Yes | No | No | No |
| PHD[d] Backonja et al [19,20] | EHR[e] Enhancements | Collection of structured data using standards and protocols | Multiple | No | Yes | Yes | No |
| Ralston et al [21] | Self-management | Web-based Interactive EHR | Diabetes | No | No | Yes | No |
| PerKApp [22] | Health promotion | Semantic inference and knowledge representation | Diabetes | No | No | No | No |
| PHKG[f] [23,24] | Health promotion | Semantic inference and knowledge representation | Diabetes | No | No | No | No |
| kHealth Sheth et al [25] | Early warning decision support system | Declarative knowledge-based reasoning and machine learning | Asthma | No | No | No | No |
| Seneviratne et al [26] | Disease characterization | Semantic inference and knowledge representation | Breast cancer | No | No | No | No |
| Chari et al [27] | Treatment recommendations | Knowledge integration | Diabetes | No | No | No | No |

[a]PHL: personal health library.
[b]ODLs: observations of daily living.
[c]SDoH: social determinants of health.
[d]PHD: Project HealthDesign.
[e]EHR: electronic health record.
[f]PHKG: personal health knowledge graph.

### Efforts to Enhance EHRs

There have been efforts to move EHRs from being mere data stores to being platforms that provide actionable information to patients, their caregivers, and health care providers. The Project HealthDesign program introduced ODLs as a major component of such enhanced EHR platforms [5,19,20]. Ralston et al describe a web-based disease self-management platform based on an interactive EHR that incorporates a diabetes module [21].

### Applications for Promoting Healthy Behavior

Some standalone applications utilize semantic inference and knowledge representation capabilities for promoting healthy behavior. For example, the PerKApp [22] leverages augmented domain knowledge in ontologies as well as reasoning rules to implement a persuasive platform targeted toward health promotion in the workplace that monitors employees' dietary and physical activity habits and sends interactive messages to persuade them to change their behaviors. A few studies leverage the notion of a personal health knowledge graph (PHKG) for patients that enables them to monitor and self-manage their chronic diseases while incorporating their ODLs [23,24]. They utilize knowledge representation to define ontologies and perform intelligence tasks on top of Resource Description Framework (RDF) graphs.

### Knowledge Integration for Disease Characterization

Some researchers have combined statistical and machine learning approaches with knowledge representation approaches. For example, Sheth et al proposed a knowledge-enabled approach to health data analytics that combines declarative knowledge-based models with probabilistic machine learning models [25]. They developed the kHealth project and deployed it as an mHealth app for decision support in patients with asthma. Seneviratne et al developed a semantic end-to-end prototype for cancer characterization [26]. Their tool utilizes a cancer staging ontology to aid physicians to quickly stage a new patient and identify risks, treatment options, and monitoring plans. Physicians can also restage existing patients or patient populations, allowing them to find patients whose stage has changed within a given patient cohort. They applied knowledge integration by converting a patient's EHR to an RDF knowledge graph and perform deductive reasoning to infer the stage of a tumor.

### Applications for Treatment Recommendations Based on Cohort Characteristics

Chari et al utilized semantic technologies and knowledge graphs to implement a tool that allows users to quickly derive clinically relevant inferences about study populations [27]. They developed a prototype workflow that utilizes an ontology to expose population descriptions in research studies through visual aids. Their goal is to enable physicians to better understand the applicability and generalizability of treatment recommendations within clinical practice guidelines.

## Innovative Technologies Used in The PHL Implementation

The PHL leverages several innovative technologies that were inspired by the requirements that we identified in the literature. We highlight some of the relevant efforts undertaken over time, including protocol specifications, vocabularies, standards, and technologies to support those requirements. We also introduce the terminology used throughout the paper, and in the results section we include several code snippets to illustrate these technologies.

### *Linked Open Data*

An abundance of scientific evidence and open data sets are available on the Web in different formats. To enhance *discoverability* and *linkability* by enabling both humans and machines to access such data, Berners-Lee proposed the *Linked Open Data* (LOD) project, which aimed to make open data available on the Web as linked data (eg, Bio2RDF) [28]. LOD is a way of connecting resources located throughout the Web by establishing a URI for each piece of data and explicitly stating how they are related to one another. LOD has led the research community to transform life sciences data sets into semantic format and make them available on the Web. LinkedCT, for example, is a ClinicalTrials.gov LOD data set that defines concepts related to diseases and interventions. The *Linked Open Research* (LOR) project leverages the LOD principles by providing an infrastructure to semantically represent research artifacts and to connect resources and the activity around them using notifications and visualizations to facilitate scientometric studies and decision making. By leveraging LOD, the proposed PHL simplifies the process of systematically and dynamically adding typed data relating to unique health and nonhealth concepts (eg, ODLs and SDoH) to the patients' digital health profiles, thereby *reducing the effort* needed for both patients and health care professionals to collect data and understand it, respectively. In addition, by leveraging the LOR initiative, the PHL enables physicians and patients to conduct scientific activities by combining global scientific knowledge discovered on the Web with local knowledge stored in their own library.

### *Web Annotation Specification*

Annotations are typically used to convey information about a resource or associations between resources. The *Web Annotation* specification [29] describes a structured data model to enable annotations to be shared and reused across different platforms. It provides a specific format for creating annotations and consuming them based on a conceptual model and a set vocabulary of terms that accommodate a certain use case. One research challenge is to explore the potentially large number of annotations to discover patterns that capture semantic knowledge not only about individual nodes and their connections but also about groups of related nodes. For example, annotating clinical trials to look for patterns is an active research area [30], and there are open data sets that can be used for that purpose. Therefore, there is a need for more automatic tools to support scientists in pattern discovery, including link predictions or discovering complex patterns of annotation (eg across multiple disease conditions and drug interventions). By leveraging LOD, the PHL enables the mining of data sets that are semantically annotated with controlled vocabulary terms and concepts (eg, risk factors) and properties (eg, risk factors associated with a disease) encoded in ontologies. Through annotations, the PHL enables a user (eg, a physician) to convey information about a resource or associations between resources (eg, a tag on a lab test or image or a comment on a blog post about a research article). Annotations also help them capture scientific knowledge and use it as a basis for conducting focused literature reviews or planning new clinical trials.

### *Representational State Transfer*

Representational state transfer (REST) is an architectural design pattern [31] for client-server communication that is centered around the following principles. First, each piece of content on the Web (both data and functionality) is considered a resource with a unique URI that provides a global addressing space for resource and service discovery. Second, resources are considered documents acted upon by Web application programming interface (API) operations (GET, POST, UPDATE, and DELETE) to manipulate those resources using HTTP as a communication protocol. Third, resources are decoupled from their representations, so their content can be accessed in a variety of formats (HTML, XML, JSON, plaintext, JPEG, PDF, etc). Finally, Web content should be designed as a network of resources that link to each other following the Hypermedia as the Engine of Application State principle. This principle enables discoverability using hypermedia controls that indicate to the resource requester a set of actions that are available to them on that resource as well as the URLs on which those actions can be performed. APIs using RESTful architecture have been widely adopted for software-to-software communication across heterogeneous distributed environments. RDF, the model driving the Linked Data Platform (LDP), follows the REST principles of identifying resources by URIs, which facilitates managing resources via HTTP operations on their URIs. It also enables hypermedia-based discovery [32]. We follow this approach for adding, deleting, and updating resources in the PHL. However, our approach solves the lack of substitutability with non-native RESTful APIs in current EHR implementations, which often hinders systems programmed for a specific API task (eg, adding a resource to Cerner at Hospital 1) from performing that same task with another incompatible API (eg, adding that same resource to Cerner at Hospital 2). Solid adopts a pattern-based approach to API design that enables applications to be compatible with APIs beyond those for which they were explicitly programmed [33].

### *Web-Scale Semantic Querying*

The LOD stack (RDF, URIs, and SPARQL [SPARQL Protocol and RDF Query Language]) makes any piece of data accessible and queryable on the Web. SPARQL can be used to execute federated queries across many endpoints on the Web. The most straightforward technique for accessing LOD data is following a URL of an RDF document through a process called *dereferencing*, which involves using the HTTP protocol to retrieve a representation of a resource identified by a URL. SPARQL endpoints offer interfaces that permit selection of data in a granular way with the ability to perform complex data retrieval from multiple *personal online data stores* (PODs) via link-following SPARQL. Solid exposes data in a document-oriented way (RDF) and provides a uniform interface to query this data.

### *Federated Linked Data Querying*

Federated queries are used to achieve web-scale integration and interoperability. Executing federated Linked Data queries on the Web requires accessing multiple data sources, which involves the discovery of data sources and determining relevant ones. Researchers have proposed discovery approaches for Linked Data interfaces based on hypermedia links and controls and applied them to federated query execution with Triple Pattern Fragments [34]. SPARQL endpoints are expensive for the server and not always available for all data sets. Downloadable dumps are expensive for clients and do not allow live querying on the web. The Linked Data Fragments framework enables client-side SPARQL querying of live Linked Data on the Web and federated querying through a triple-pattern interface, providing a much faster, less expensive solution [34].

## *Methods*

We conducted a thematic assessment of patients' requirements and used the identified requirements to develop a use case scenario that motivates the need for a PHL. We explain the technological infrastructures used to build the PHL platform, including the Solid platform and knowledge graphs.

### Thematic Assessment of Patient Requirements

The patient requirements for a PHL that we identified in the literature [7-12] fall into three broad themes: (1) construction and management of the library, (2) dynamic discovery and integration of new knowledge related to data and types stored in the library, and (3) the ability to leverage the knowledge stored in the library through digital interventions (Table 2). We assessed the technological innovations required to meet those requirements. To address R1.2 and R3.2, we need an infrastructure that supports privacy by design. To enable patients to effectively share data and knowledge, we need a platform that supports sharing not only with individuals (R5, R11) but also with organizations (R6) and devices (R10, R13.2). To enable patients to selectively define and store *types* of knowledge or data (R1.3, R3.1), we need to leverage Semantic Web technologies. Finally, to incorporate dynamic discovery and knowledge enrichment, we need to store patient's data using a KG (R4.1, R4.2) in a machine-readable format. Besides these functional requirements, the platform should support nonfunctional requirements, including security, integration, and interoperability. Our innovative technologies meet all of these requirements.

Table 2. Some requirements for a PHL (from a patient perspective, per the literature).

| Requirement | Description |
| --- | --- |
| **I. Construction and management of digital health state** | |
| R1.1 Integration | *Construct* a PHL[a] by *bringing* a patient's *data together* in a *trustworthy*, *usable*, and *useful* library by gathering different *types* of *knowledge* into a *single resource* |
| R1.2 Security and privacy | |
| R1.3 Semantic technologies: ontologies | |
| **R2 Management:** | *Manage* the PHL by *creating*, *reading*, *updating*, or *deleting* resources |
| a. RESTful[b] resources | |
| b. CRUD[c] operations | |
| R3.1 Semantic technologies: ontologies | *Decide* what *types* of data should be kept and who *has access* to that data |
| **R3.2 Privacy:** | |
| a. What: resource | |
| b. Who: agent | |
| **II. Dynamic knowledge discovery and integration** | |
| R4.1 Dynamic knowledge discovery | *Seek* health data from *constantly changing public sources*, enriched with *new streams* and *types* of *data* |
| R4.2 Knowledge enrichment | |
| R5 Knowledge sharing with individuals | Decide what *types* of data are important to *collect*, *manage*, and *share* |
| R6 Knowledge sharing with organizations | *Share* data with *citizen science* and *research initiatives* |
| **III. Processing digital health state (digital interventions/mHealth[d] apps)** | |
| **Interaction-based usage** | |
| R7.1 Searching | *Search* through the PHL using *intelligent mapping* for *vocabulary* used to describe *resources* in the patients' *profiles* |
| **R7.2 Semantic technologies:** | |
| a. Unique resource representation | |
| b. Vocabulary mapping | |
| R8.1 Semantic technologies: annotations | *Annotate* results from patient's participation in clinical trials to *look for patterns* |
| R8.2 AI[e]: pattern detection | |
| **Notification-based usage** | |
| R9 Dynamic knowledge discovery | Receive *alerts* about *new* data *related* to *topics* covered in their PHL |
| R10 Wearable device agents (ODL[f] data) | Play an active role in staying healthy by *monitoring* their *progress* |
| R11 Knowledge sharing with individuals | *Stay current* with *treatment options* and *clinical trials* for a *family member* with a debilitating condition |
| **R12 Semantic technologies:** | *Find* and *use* information including *text summarization*, *knowledge mapping*, etc |
| a. Text summarization | |
| b. Knowledge mapping | |
| **R13.1 Intelligent mHealth apps:** | Access *digital assistance via personalized alerts* and suggestions, text summarization, *literacy* aids, *translations*, etc |
| a. Digital assistants | |
| b. Recommender systems | |
| R13.2 Software agents | |

[a]PHL: personal health library.
[b]REST: representational state transfer.
[c]CRUD: creating, reading, updating, or deleting.
[d]mHealth: mobile health.
[e]AI: artificial intelligence.
[f]ODL: observations of daily living.

## Scenario

We present a scenario that involves Bob as the main actor and Alice and Mary as supporting actors to demonstrate how a diabetic patient can benefit from the PHL and use the mHealth app for diabetes self-management (Textbox 1).

**Textbox 1.** Use case scenario of self-management for a diabetic patient.

> *Bob* is an African American adult with diabetes equipped with a smartwatch and smartphone that collect physiological data (eg, step counts) in real time. The social app on his smartphone queries his decentralized personal health library (PHL) to deliver tailored push notifications to support behavior change related to chronic disease self-care. Depending on sensor readings and other information in his PHL, the app provides personalized and tailored recommendations for healthy eating, physical activity, medication taking, and/or visiting health care providers. The health recommendations also take into account different characteristics of the nearby points-of-interest. For example, as Bob has another comorbidity (asthma), running activities must be avoided.
>
> *Alice* is a patient who is under treatment for cancer and is also part of Bob's social network. She would like to report acute health conditions and side effects of using medications by sharing a notepad with her physician.
>
> *Mary* is a physician who follows up with both Bob and Alice. Through the PHL, she would like to interact with her patients and to be able to access their test results easily. She would also like to conduct research about their health conditions using the content generated through their PHLs and her own by tracking publications and scientific observations from trusted public knowledge sources.
>
> *A clinic or lab* would like to access the PHLs of Alice and Bob to share the results of their lab tests and to follow up on the test and visit history. It also wants to share those tests with their physician, Mary.

In the following, we describe two of the main technological infrastructures used to construct, manage, and integrate the types of knowledge stored in the PHL. Namely, we describe the Solid platform and the role of personal knowledge graphs in designing the PHL and the mHealth app.

### *Social Linked Data (Solid)*

To extend the current functionality of the Web (World Wide Web Consortium [W3C] standards and protocols) by applying LOD principles, Berners-Lee proposed the Solid project [35,36]. Solid is shaping the future vision of the decentralized Web (Web 3.0) by enhancing the technologies used to build Web 2.0 while bringing back the privacy and freedom-oriented values of Web 1.0. Solid uses the *WebID specification* [37] to implement a global identity management architecture based on the notion of decentralized identity providers. Coupled with the WebID-TLS decentralized authentication protocol, a WebID enables a global web-scale single sign-on. Moreover, Solid follows a unique architecture for building applications by separating users' data from the applications that use that data, which guarantees not only privacy and true ownership but also flexibility. Users can store their data among several Solid-compatible PODs hosted on Solid-enabled Web servers (Figure 1) that users can either install on their own machines or obtain from a listed provider and selectively authenticate applications to access and process specific resources within those servers. When users register for an identity, their WebID profile document and associated cryptographic key is stored on their main POD server (Figure 1). By leveraging Solid, the underlying storage for patients' digital data in the proposed PHL can be implemented in several ways, for example, file systems, key-value stores, and relational or graph database systems (Figure 1). In addition, accessing physical data and the metadata will be performed through an allocated semantic layer so that changing or reorganizing the data sources does not cause interruption in the application. Solid is a stack of protocols and standards, so any mHealth app that utilizes the PHL will enable users to maintain control of their data so long as the app is built in a Solid-compatible way.

### *Personal Knowledge Graphs*

A personal knowledge graph (PKG) [38] provides a new research frontier toward building intelligent applications. Assembling data from distributed sources is often challenging, but by leveraging the LOD stack (RDF, URIs, and SPARQL), our PHL platform can achieve such assembly by building an RDF representation of a PKG for each patient that maintains a historical representation of that patient's digital health state. Applications can query those graphs to render different aspects through visual aids and aggregate data by accessing the PODs of the target patient, PODs belonging to other users, and external Web resources (Figure 1).

## Results

### Summary

We present the results of assessing the requirements in the previous scenario, the prototype of the PHL and mHealth app, and our evaluation plan. We follow the three main themes identified in Table 2 to demonstrate how the different innovative technologies incorporated into the PHL meet these requirements through actions performed by the actors in our scenario (summarized in Textbox 1). In Textbox 2, we distill the most important requirements that the three actors and the involved organization or organizations will need in the proposed PHL and show how we leverage Solid and PKGs, among other technologies, to achieve those requirements.

**Textbox 2.** Requirements identified in the use case scenario from Textbox 1.

---

**Functional requirements:**

*Agents*

- Device agents (smartwatch, smartphone)
- Person agents (Mary/physician, Alice/another patient)
- Organization agents (clinic)
- Software agents (blogging app, calendar app)

*Integrate different types of knowledge from different sources*

- Registries: comorbidity (asthma)
- Global knowledge: blogs (HTML), articles (PDF)
- SDoH: neighborhood with low walkability score
- ODL: healthy eating, physical activity, medication taking, and visiting health care providers

*Usage patterns*

- Annotation of research articles and clinical trials
- Sharing knowledge and resources

**Nonfunctional requirements:**

*Security, privacy, interoperability*

- Clinic and PHL software integration (posting test results)
- PHL and EHR software integration (patient's self-reported outcome)

---

## Construction and Management of the PHL

We describe how the three actors in our scenario (Textboxes 1 and 2) can use our platform to construct their PHLs (Figure 2). This includes decentralized identity through WebIDs, main and extended profile documents (Figure 2, F1-F3), trusted agents (Figure 2, F4), and resource management (Figure 2, F5).

**Figure 2.** Main PHL features that meet some of the patients' requirements (Table 2) demonstrated through the PODs of Alice, Bob, and Mary. Bob's POD contains his main profile document in RDF-based KG representation. Social interactions within the PHL ecosystem include: (1) Alice and Mary can subscribe to Bob's channel using their WebIDs. (2) Alice can share her lab tests by pushing them to Mary's inbox. (3.1) Alice can share a notepad with Mary to discuss her lab results. (3.2) Alice can add annotations or comments to message content in Bob's diabetes channel. (4) Software from a clinic or other provider can share test results with Alice by performing a POST Web API operation on the unique URI of her inbox. POD: personal online data store.

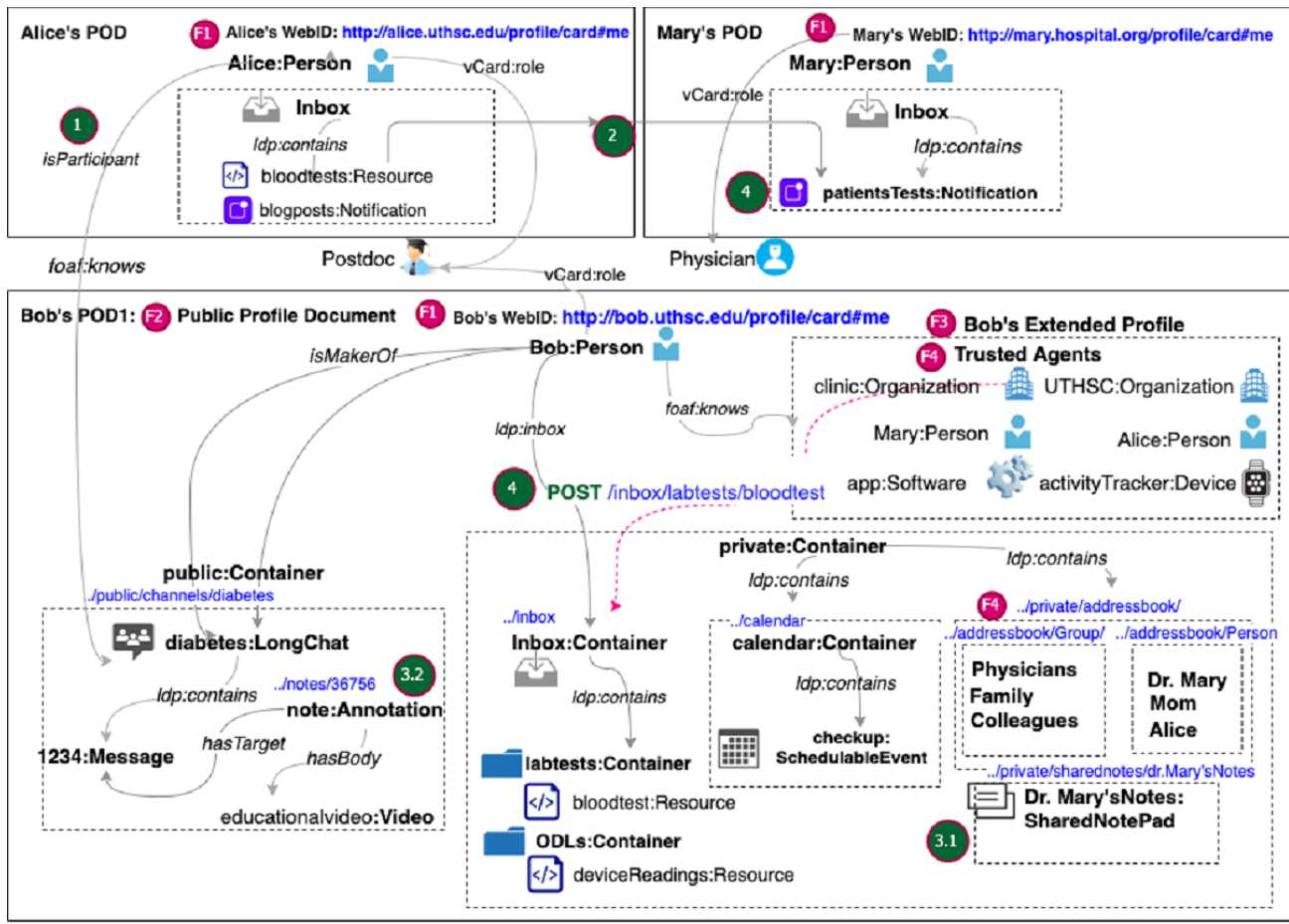

## Decentralized Identity Through WebIDs

First, the three actors generate unique WebIDs (Figure 2, F1) to securely log in to their main RDF-based PHL *profile document* (Figure 2, F2). Once they set up their main PHL profiles, they can build a *Web of Trust* using the FOAF (Friend of a Friend) vocabulary (eg, foaf:knows) by linking their main profiles to one or more extended profiles (Figure 2, F3). In their extended profiles, our actors can keep lists of vCard URIs of trusted agents and selectively grant them access to content in their PHL. For example, Bob can create an extended document in which he stores his friends list (Figures 3 and 4). Once they set up their profiles, users can manage content within those documents. In the following section, we explain how Bob manages content under his PHL and how he adds trusted agents.

**Figure 3.** Bob's main PHL profile document with a reference to his friends' extended profiles.

```
@prefix foaf: <http://xmlns.com/foaf/0.1/> .
@prefix rdfs: <http://www.w3.org/2000/01/rdf-schema#> .
<> PersonalProfileDocument ;
<#me> a: Person ;
   :name "Bob";
   rdfs:seeAlso https://bob.uthsc.edu/friends     .
```

**Figure 4.** Bob's extended profile document (friends) that identifies Alice and Mary as trusted agents by establishing foaf:knows relations with their WebIDs.

```
@prefix foaf: <http://xmlns.com/foaf/0.1/>.
<> a foaf:PersonalProfileDocument ;
  foaf:maker <https://bob.uthsc.edu/profile#me> ;
  ..
  foaf:knows <https://uthsc.edu/p/Alice#MSc> ;
  foaf:knows <https://hospital.org/people/Mary/card#me> .
```

### Hierarchical Resource Representation

Whether it is a person, an inbox, a file, an image, a notification, or a relationship, content within the PHL is represented as a collection of Web resources. Patients can organize resources in their PHL as a hierarchy of nested *containers*. For example, an event is nested inside a calendar and a message is nested within a chat channel. Both containers and resources conform to the LDP BasicContainer specification. The different actors in our scenario can start with default containers provided by the PHL.

For example, the inbox container gets created in the PHL as a default container preconfigured with live notifications. The patient's inbox is also discoverable through the ldp:inbox property specified in the Linked Data Notification (LDN) specification [39]. Beyond default containers, patients can define their own resources or containers. For example, Bob can add the calendar and diabetes folders under his public or private folders (Figure 5). He can also set up a chat channel about diabetes as a resource of type LongChat and nest that within the Diabetes folder (Figure 2, F5).

**Figure 5.** Hierarchy of containers under Bob's PHL.

```
@prefix ldp:<http://www.w3.org/ns/ldp#>.
    <ldp:BasicContainer> ;
        ldp:contains <calendar>, <diabetes>, <inbox> .
<calendar>
    A <..#Event>.

< diabetes>
    a <https://www.w3c.org/ns/type#LongChat> .
<inbox>
    a <https://www.w3c.org/ns/ldp#inbox> .
```

### Flexible Data Representations

The PHL supports reading and writing resources in different formats: (1) structured Linked Data resources (eg, RDF, HTML+RDFa (RDF in Attributes), etc), (2) binary data (eg, images, videos, webpages), and (3) non–linked data structured text. While the PHL enables building applications with nonlinked resources, using RDF-based linked data provides extra benefits in terms of interoperability with the rest of the ecosystem.

### Trusted Agents

Agents can be persons, organizations, devices, or software. For example, Bob can add the clinic as an organization agent, Alice and Mary as person agents, his mobile phone as a device agent, and the diabetes self-management mobile app as a software agent (Figure 2, F4). Access can be granted either to individual agents or agent groups. For example, Bob can define two work-groups (Physicians and Caregivers) in an extended document (work-groups) (Figure 6). In that document, he can list Mary and Alice as members using their WebIDs. He can then grant each of these agent groups fine-grained access permissions to his shared-notepad resource. Bob can reference a group as a resource under the work-groups document (Figure 7).

In addition, since he is interested in tracking and gathering data about diabetes from a blogging app, he can add the app under the trusted apps section of his extended profile document (Figure 8).

**Figure 6.** Bob's work-groups document that defines Physicians and Caregivers as groups.

```
@prefix   vcard:   <http://www.w3.org/2006/vcard/ns#>.
<#Physicians>
      a                vcard:Group;
      vcard:hasUID     <urn:uuid:1234:ABGroup>;
      # Physicians group members:
      vcard:hasMember  <https://Hospital.org/profile/card#me>;
<#Colleagues>
      a                vcard:Group;
      vcard:hasUID     <urn:uuid:5678:ABGroup>;
      # Caregivers group members:
      vcard:hasMember  <https://alice.uthsc.edu/profile/card#me>.
```

**Figure 7.** Individual and Group authorizations to Bob's notepad.

```
@prefix   acl:   <http://www.w3.org/ns/auth/acl#>.
# Individual authorization - Bob grants Mary Read/Write/Control access on his notepad
<#authorization1>
      a             acl:Authorization;
      acl:accessTo  <https://bob.uthsc.edu/data/shared-notepad>;
      acl:mode      acl:Read,
                    acl:Write,
                    acl:Control;
      acl:agent     <https://hospital.org/profile/card#me>.
# Group authorization - Bob grants members of Physicians group Read/Write access to
his notepad
<#authorization2>
      a             acl:Authorization;
      acl:accessTo  <https://bob.uthsc.edu/data/shared-notepad>;
      acl:mode      acl:Read,
                    acl:Write;
      acl:agentGroup <https://bob.uthsc.edu/work-groups#Physicians>;
```

**Figure 8.** Bob's trusted apps.

```
<#me> acl:trustedApp [ acl:origin  <https://calendar.app.com>;
                      acl:mode    acl:Read,
                                  acl:Append].
<#me> acl:trustedApp [ acl:origin  <https://blogging.app.com>;
                      acl:mode    acl:Read,
                                  acl:Write].
<#me> acl:trustedApp [ acl:origin  <https://diabetesSelfManagement.app.com>;
                      acl:mode    acl:Read,
                                  acl:Write].
```

*Access Control Lists*

The PHL uses the W3C Web Access Control (WAC) ontology to describe Read, Write, Control, and Append access control modes (eg, acl:mode acl:Read, Figure 9) at the level of a container or resource. Each resource can have an associated access control list (ACL) resource. If a container or resource does not have an ACL, it inherits the authorization of its parent container. For example, the default ACL on the inbox container is append-only by the public. Bob can associate the lab_test resource within his inbox with a corresponding ACL resource (lab_test.acl). Then, within that ACL resource, he can specify trusted agents and their corresponding access modes. For example, he can limit access to his friends' list extended profile document (Figure 4) by defining an ACL rule using the WAC ontology (Figure 9).

**Figure 9.** An ACL rule granting Read permission to Alice and Mary on Bob's "Friends" document.

```
@prefix acl: <http://www.w3.org/ns/auth/acl#> .
<#FriendsOnly>;
    acl:accessTo <https://bob.uthsc.edu/friends> ;
    acl:agent <http://uthsc.edu/people/Alice#Msc>,
              <http://hospital.org/people/Mary/card#me> ;
    acl:mode acl:Read .
```

## Dynamic Knowledge Discovery and Integration

In this section, we describe how dynamic discovery and integration can be achieved through linkability, the ability to interact with the PHL content through annotations, rich embedding, and social interactions between the different actors in our scenario.

### Linkability

Resources generated by each of the three actors get stored in their PHLs, with the possibility of linking resources in their PHLs to resources in other users' PHLs (Figure 2, F7). For example, if Alice comments on a message stored under the diabetes channel resource in Bob's PHL, her message will be stored under her PHL but links to Bob's message using the hasTarget Link type defined in the Web Annotation Ontology (WAO) (Figure 10).

**Figure 10.** Alice's comment is linked to Bob's message using the hasTarget link type of the Web Annotation Ontology.

```
@prefix wao: <http://www.w3.org/ns/oa#>.
<https://alice.uthsc.edu/comments/36756>
    wao:hasTarget
        <https://bob.uthsc.edu/messages/diabetes/1234>
```

### Annotations, Rich Embedding, and Social Interaction

The PHL can be used as a decentralized authoring, annotation, and social interaction framework that can be accessed from several platforms (eg, Web browsers and mobile apps). By implementing the Web Annotation specification [29], the PHL enables physicians to annotate content within resources stored under their PHLs. As a physician-researcher, Mary can benefit from the content generated through her PHL about diabetes. To this end, she can announce her inbox so she can receive LDN notifications of scholarly activities related to diabetes (eg, published articles, annotations in peer reviews [29], scientific observations). She can add identifiers to important concepts within her received content (eg, article) and add descriptive markup to those identified concepts. Her PHL uses the identifiers to automatically generate URIs for every article section to make it easy for others to refer to them or link them to external knowledge resources. Using Mary's descriptions, the PHL automatically generates RDFa markup documents, and by implementing the protocol, it enables her to expose those generated documents as Linked Data for other researchers to consume.

The PHL also supports rich embedding, whereby a researcher or a physician can embed raw data within a document in several formats and add provenance links (eg, nanopublications). For example, having Mary as a physician in the chat channel can add credibility to the discussion and enrich the knowledge exchanged. She can highlight certain statements exchanged in a message and correct a misconception. She can also comment on shared knowledge in a particular message and probably refer patients to more trusted sources of knowledge (eg, Centers for Disease Control and Prevention, World Health Organization). Similarly, when Bob receives information about treatment options from Mary as his physician, he can integrate it with information obtained from the blogging app. For example, he can highlight concepts in a blog post relating to diabetes and related interventions and link them to the same concepts in Mary's shared notepad or concepts shared by other users in his diabetes channel.

Mary, Bob, and Alice can perform social interactions within the PHL ecosystem in different ways (Figure 2). For example, (1) Alice and Mary can subscribe to Bob's channel using their WebIDs; (2) Alice can share the results of her lab tests by pushing them to Mary's inbox; (3.1) Alice can share a notepad with Mary to discuss digital health knowledge that she obtained from other sources, and Mary can interact with the content by adding annotations, refining provenance links, or linking words to scientific concepts; (3.2) Alice can also add annotations or comments to message content in Bob's diabetes channel; and (4) software from a clinic or other provider can share test results with Alice. Users get notifications for every activity performed on content under their PHLs, including annotations, replies, shares, reviews, citations, links, bookmarks, and even likes.

### Data Access Through RESTful HTTP Operations

Data stored in the PHL is managed in a RESTful way; new resources are created under a container by sending them to the container's unique URI via an HTTP POST operation. The PHL supports two-way server-to-server and client-to-server communication. Either way, the requesting agent performs an HTTP operation on a given resource URI under a given POD server. Sending lab results from the clinic to Alice's inbox is an example of a server-to-server communication. The software agent (eg, Cerner) hosted on the clinic's server performs a POST operation on the /inbox/lab-tests resource on the POD server hosting Alice's PHL (Figure 2, step 4). A self-reported outcome is an example of a client-to-server communication that originates from a patient's PHL to external software. For example, Bob can report an allergy, which generates a POST request on a URI under the server hosting his EHR.

Agents do not need to know the internal structure of a patient's PHL. Each resource in PHL is its own SPARQL endpoint, which can be advertised through Link headers that can be discovered by sending HTTP GET/HEAD operations on the main PHL URI. Performing a GET operation on any URI that ends with a * returns an aggregate of all the resources that match the indicated pattern. For example, to fetch all diabetes self-management recommendations under Bob's diabetes container, an agent can send a single GET request to the /data/diabetes/* resource on the server hosting his PHL profile (Figure 11).

**Figure 11.** A GET request on Bob's diabetes messages folder under his POD.

```
REQUEST:
GET /data/diabetes/* HTTP/1.1
Host: https://bob.uthsc.edu
RESPONSE:
HTTP/1.1 200 OK
<>
    a <http://www.w3.org/ns/ldp#BasicContainer> ;
    <http://www.w3.org/ns/ldp#contains> <diets>, <exercises> . <medications>
<diets>
    a <https://bob.uthsc.edu/ns/type#Message> .
<exercises>
    a <https://bob.uthsc.edu/ns/type#Message> .
<medications>
    a <https://bob.uthsc.edu/ns/type#Message> .
```

Application-to-server communication happens when an application pushes content to a user's PHL. Users can interact with their PHLs through different apps installed on their phones after signing in using their WebIDs. For instance, when Bob installs the diabetes mHealth app on his phone, recommendations are pushed to the corresponding container within his messaging POD (Figure 1). By signing up, his PHL sends periodic GET requests to the app to get the latest recommendations. Similarly, Mary can add follow-up events using the calendar app of her choice, and that app can share those events by pushing them as resources under Bob's calendar container by sending POST requests on that container's URI.

### *Prototype Design of the PHL and mHealth App*

Figure 12 shows the main interface design of the mobile app and Figure 13 shows the PHL. Through the PHL, patients can set their preferences to tailor the *content* in the desired recommendations in terms of focus (eg, diet, exercise, and medication adherence), frame (eg, educational, motivational, goal-based), and frequency (daily, weekly, etc).

**Figure 12.** Mobile app for chronic disease self-management.

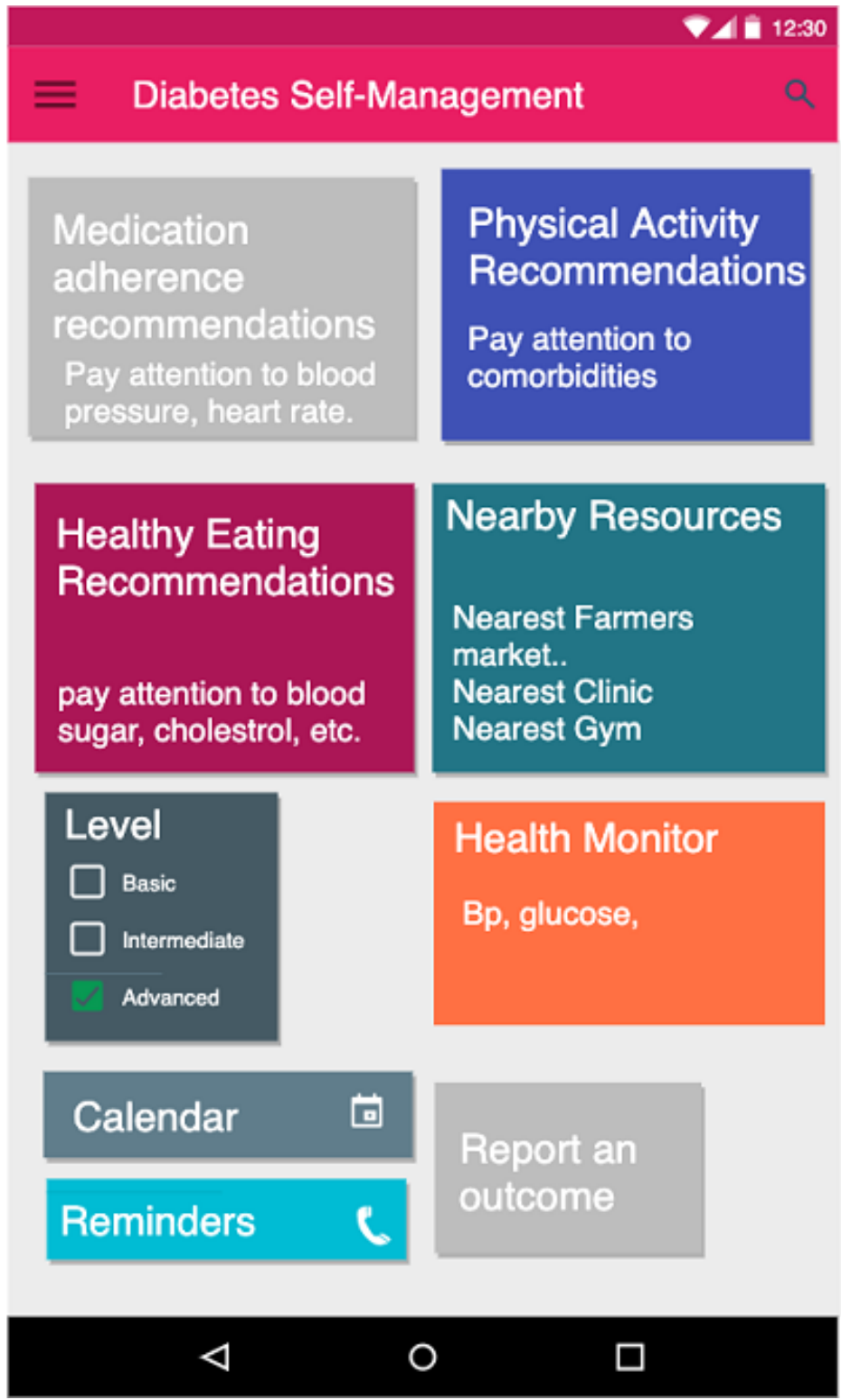

**Figure 13.** The PHL that enables the app in Figure 12. Through the PHL patients can perform (a) Chronic disease self-management, obtain (b) External knowledge and resource suggestions, and (c) manage trusted agents. The PHL can personalize recommendations by utilizing knowledge about monitored ODL readings, location-based detection of SDoH, external knowledge suggestions, and EHRs. Other features include reminders of medications and shared notepads with physicians.

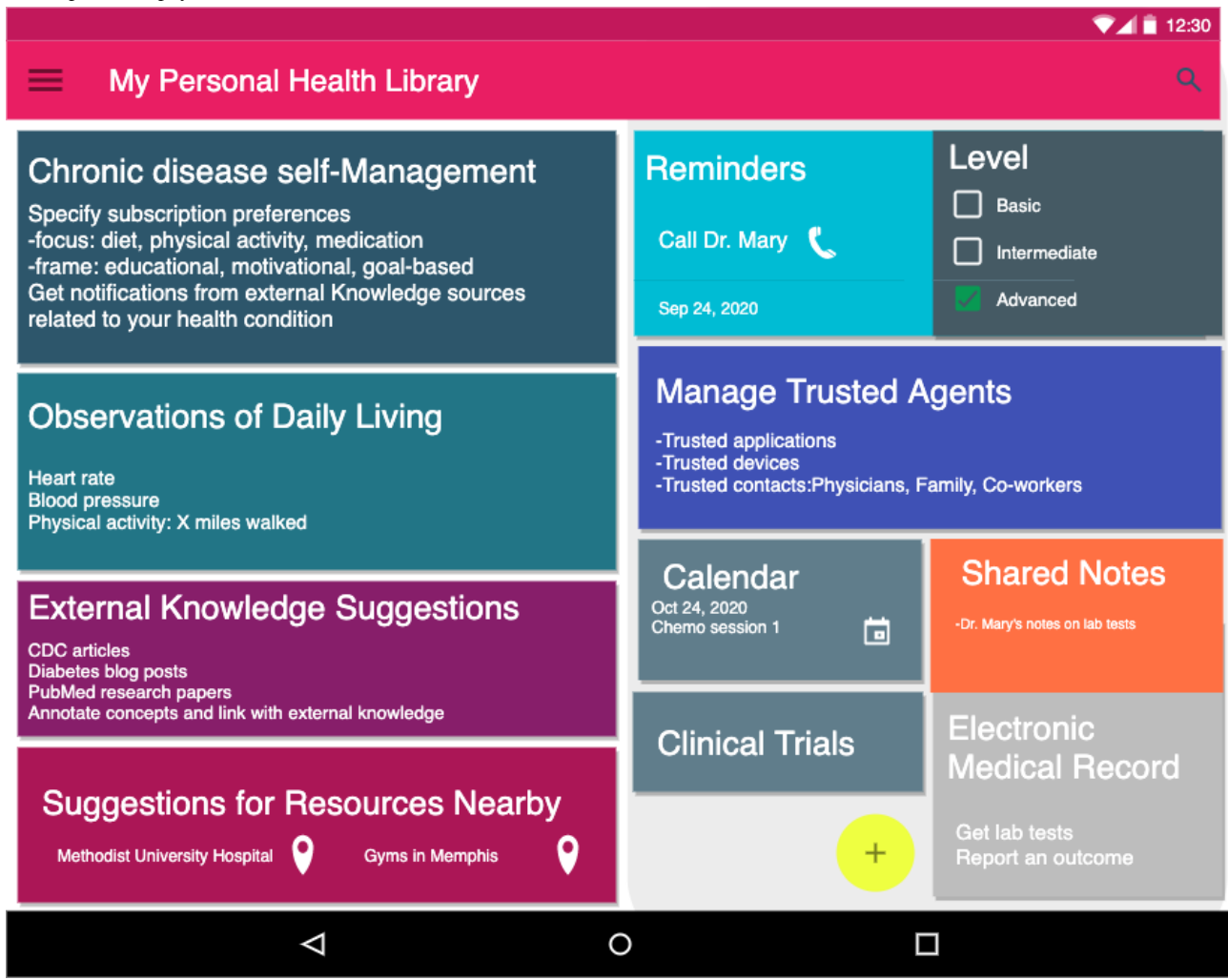

The app, on the other hand, sends enriched message recommendations based on patients' preferences. By accessing dynamic knowledge in the PHL, the app can provide real-time hybrid recommendations that are both *content-* and *context-*based. To capture context, the PHL collects both ODL and SDoH data from activity trackers and population-level neighborhood characteristics. It utilizes both the data and the semantic conceptual hierarchies of knowledge types stored in the library to infer new knowledge and collect more evidence. For example, if a patient lives in a zip code that has a low walkability score, the app avoids sending recommendations that encourage walking in the neighborhood. In addition, if their device reading shows a high heart rate or if their EHR shows they have asthma, the app avoids recommendations of physical activities that would severely affect their health conditions. In addition to textual message recommendations, the app provides resource suggestions within the patients' zip code area or provides language-speaking services that respect their race or ethnicity.

### Sources of Data and Knowledge

The PHL utilizes and integrates data and knowledge from several sources:

1. Historical data collected systematically through regional registries—in the case of the mobile app proposed in this paper, we utilize the Diabetes Wellness and Prevention Coalition (DWPC) regional registry, which aims to improve care for people with obesity, diabetes, and other obesity-associated chronic conditions. It provides clinical information, including hospital and clinic visits, as well as labs, diagnoses, and medications.

2. ODL data collected through mobile devices and activity trackers, such as physical exercises, heart rate, etc

3. Population-level SDoH data, such as walkability scores in a neighborhood and access to public transport, among others

4. Public knowledge obtained from research and news articles, blogs, and social networks

### Enabling Technologies for Building the mHealth App

To implement PHL-enabled mobile apps, we will utilize a RESTful API that will expose the RDF-based PKG data model stored in a patient's PHL. We will leverage the Solid technology stack from within JavaScript-based environments, which will enable us to integrate data by invoking APIs exposed by different POD servers. For Linked Data manipulation and querying, we will utilize RDFlib and LDflex. We will use the Solid-authenticated RDFlib API and query engine for advanced parsing and querying of the patient's RDF-based PKG data model stored in a patient's PHL.

In addition to querying the patient's local PKG data model, we will utilize the LDflex domain-specific language to query any Linked Data resource on the Web, which enables dynamic external knowledge discovery and integration. LDflex provides concise expressions that allow us to perform complex federated query execution without having to craft all HTTP requests required in a query and without hardcoding resource URLs. We provided the query engine with the root node (eg, Bob's WebID), an entry point (eg, Bob's inbox/lab_tests/test1), and a property (eg, testResult) to query Bob's health knowledge graph data model. The engine uses the entry point to recognize the context in which the query is executed and resolves the expression into an actual path on Bob's graph. Expression execution involves several steps: obtaining Bob's WebID URL, resolving the terms included in the expressions (eg, lab_test) to their URLs, creating the SPARQL query that represents the expression, fetching the document of the root user (Bob) through an HTTP request, and executing the SPARQL query on the document and returning the result.

**Evaluation Plan**

Before fully implementing the PHL and the mHealth app, we will conduct two evaluation studies. In this section, we provide a protocol synopsis for each study.

### Study 1: User-Centered Design and Formative Evaluation of the Prototype

The so-called "digital divide" may hinder the adoption of digital interventions, but the use of human factors engineering can overcome some of the challenges in that regard [40,41]. We will conduct a descriptive, iterative, user-centered design and formative evaluation study by seeking feedback from pre-development focus groups, specifically participants from regional hospitals, including patients, caregivers, health care professionals (clinicians, residents, physicians), and health education professionals. The goal is to reveal issues related to the (1) usability, (2) clinical content, and (3) educational content of both the (a) PHL platform and (b) recommendations produced by the app. We will organize a workshop to both verify initial requirements, scenarios, and storyboards and identify new ones. We will use our findings to refine the initial requirements and develop new scenarios and generate new storyboard simulations. We will assign participants to focus groups and run different scenarios and dashboard simulations by them. We will use the System Usability Scale and the EHR Usability Scale to gather feedback [42,43]. Finally, we will apply thematic assessment to the resulting transcripts and use the results to design the final PHL and mHealth app following the refined storyboard simulations.

### Study 2: Pragmatic Clinical Trial for Evaluation of the Digital Intervention

To evaluate the effectiveness of this intelligent digital intervention compared to existing interventions on the population of patients with diabetes, we will replicate the pragmatic randomized controlled trial design described by Bailey et al [44], with variations. The three primary *outcomes* that we will test are the number of days in the previous week that participants (1) ate healthy meals, (2) participated in at least 30 minutes of physical activity, and (3) took medications as prescribed. Participants will be adults age ≥18 years who have uncontrolled diabetes (A1c≥8), exhibit one or more additional chronic conditions seen in clinics in medically underserved areas of the mid-South, and possess a phone with texting and voicemail capability. Intervention arms will be standard motivational SMS text messaging (TM) and intelligent recommendation–enhanced TM (IR-TM), and the control arm will be usual care. We will capture the patients' perceptions of diabetes self-care activities using subscales of the revised Summary of Diabetes Self-Care Activities questionnaire administered over the study follow-up period. We will also use the DWPC registry to obtain clinical data, including A1c, body mass index, and blood pressure. We will test multiple hypotheses to determine the comparative effectiveness of the control and intervention arms for each primary outcome. Sample size and power estimates for these types of digital interventions will follow the approach described by Bailey et al [44], with the assumption that effect sizes will range from small (standardized difference=0.375) to medium (=0.50). To obtain power estimates, projected mean changes over 12-month follow-up from baseline (mean for 12-month follow-up minus mean for baseline) of the control and intervention arms for each primary outcome will be obtained using results reported in the literature. We expect the TM and IR-TM arms to have adequate power to detect meaningful changes from baseline with respect to all three primary outcome variables compared to the usual care arm will.

## Discussion

Our previous research has identified that minorities and low-income, underserved communities are disproportionately affected by chronic diseases [45] such as obesity [46], a risk factor for diabetes, heart disease, and cancer. In this paper, we describe a personal health library–enabled mHealth app that provides hybrid recommendations by incorporating SDoH and ODLs in addition to digital health information to provide insights for informing preventive digital interventions in chronic disease management.

The PHL gathers different types of knowledge into a single searchable resource. While there has been some effort to build similar systems, the novelty of the proposed approach lies in (1) providing a decentralized yet linked architecture; (2) supporting interoperability, portability, knowledge mapping, and reasoning by following protocol, format, and vocabulary standards; (3) building trust with patients by facilitating true ownership over their data and appropriate reporting; and (4) giving those patients fine-grained access control mechanisms.

The PHL will not only help patients and their caregivers to assume a central role in making decisions regarding their health but also equip health care providers with informatics tools that will support the collection, interpretation, and dissemination of the collected knowledge. By moving health care beyond clinical settings, clinicians can benefit from the PHL in leading new treatment regimens and keeping in touch with their patients between office visits.

Future work will focus on further implementation of an end-to-end framework of an intelligent recommender and digital librarian, including text summarization, knowledge mapping, and personalized resource suggestions. In achieving this goal, we will incorporate artificial intelligence techniques and knowledge representation methods that have been successfully used in our previous works [47,48]. Other ongoing tasks will include establishing a clinical trial of the app and recruiting participants to fully evaluate the app. Future work will also focus on the enrichment of patients' health knowledge graphs to improve the reasoning capabilities of the knowledge layer.

Finally, we plan to expose parts of the PHL functionality as an open service for fostering the development of third-party applications that may provide motivational technological support in several national and international projects crossing different domains of interest. To achieve this, the library will serve as an API for querying, managing, and using a patient's health RDF-based knowledge graph. This will give the community access to the infrastructure of the library to enable building applications that benefit from the library for other phenotypes.

## Conflicts of Interest

None declared.

## Abbreviations

**ACL:** access control list
**API:** application programming interface
**DWPC:** Diabetes Wellness and Prevention Coalition
**EHR:** electronic health record
**IR-TM:** intelligent recommendation–enhanced TM
**LDP:** Linked Data Platform
**LDN:** Linked Data Notifications
**LOD:** Linked Open Data
**LOR:** Linked Open Research
**ODLs:** observations of daily living
**PHKG:** personal health knowledge graph
**PHL:** personal health library
**POD:** personal online data store
**REST:** representational state transfer
**RDF:** Resource Description Framework
**RDFa:** RDF in Attributes
**SDoH:** social determinants of health
**Solid:** Social Linked Data
**SPARQL:** SPARQL Protocol and RDF Query Language
**TM:** text messaging
**W3C:** World Wide Web Consortium
**WAC:** Web Access Control
**WAO:** Web Annotation Ontology